\relax
\documentclass[twocolumn]{article} 
\usepackage{times}  
\usepackage{helvet}  
\usepackage{courier}  
\usepackage[hyphens]{url}  
\usepackage{graphicx} 
\usepackage{natbib}  
\usepackage{caption} 
\DeclareCaptionStyle{ruled}{labelfont=normalfont,labelsep=colon,strut=off} 
\usepackage{algorithm}
\usepackage{algorithmic}

\usepackage{newfloat}
\usepackage{listings}
\floatstyle{ruled}
\newfloat{listing}{tb}{lst}{}
\floatname{listing}{Listing}

\usepackage{amsmath}
\usepackage{amssymb}
\usepackage{amsfonts}
\usepackage{mathtools}
\usepackage{subfig}
\usepackage{multirow}
\usepackage{tabularx}
\usepackage{theorem}
\makeatletter
\newcommand{\thickhline}{%
    \noalign {\ifnum 0=`}\fi \hrule height 1pt
    \futurelet \reserved@a \@xhline
}
\newcolumntype{"}{@{\hskip\tabcolsep\vrule width 1pt\hskip\tabcolsep}}
\makeatother
\newtheorem{theorem}{Theorem}[section]

\usepackage{cleveref}
\usepackage{xr}
\usepackage{bbm}
\usepackage{upgreek}
\usepackage{bm}
\usepackage{enumitem}
\usepackage{array}
\usepackage{ragged2e}
\usepackage{stackengine}
\usepackage[utf8]{inputenc}
\usepackage{authblk}

\usepackage{stmaryrd}

\newcommand{\bemul}{\begin{multline*}}
\newcommand{\bee}{\begin{eqnarray*}}
\newcommand{\eee}{\end{eqnarray*}}
\newcommand{\been}[1]{\begin{eqnarray}\label{#1}}
\newcommand{\eeen}{\end{eqnarray}}
\newcommand{\began}[1]{\begin{gather}\label{#1}}
\newcommand{\bega}{\begin{gather*}}
\newcommand{\bealn}[1]{\begin{align}\label{#1}}
\newcommand{\beal}{\begin{align*}}
\newcommand{\bealatn}[2]{\begin{alignat}{#1}\label{#2}}
\newcommand{\bealat}{\begin{alignat*}}
\newcommand{\bexalatn}[1]{\begin{xalignat}\label{#1}}
\newcommand{\bexalat}{\begin{xalignat*}}

\newcommand{\mbb}{\mathbb}

\def\bx{{\mathbf x}}  
\def\by{{\mathbf y}}
\def\bz{{\mathbf z}}
\def\bA{{\mathbf A}}
\def\bB{{\mathbf B}}

\def\bI{{\mathbf I}}

\def\bX{{\mathbf X}}

\def\texitem#1{\par\smallskip\noindent\hangindent 25pt
               \hbox to 25pt {\hss #1 ~}\ignorespaces}

\newcommand{\scrP}{\mathcal{P}}

\newcommand{\scrS}{\mathcal{S}}

\newcommand{\scrZ}{\mathcal{Z}}

\newcommand{\btheta}{\boldsymbol{\theta}}
\newcommand{\bTheta}{\boldsymbol{\Theta}}

\newcommand{\bpi}{{\boldsymbol{\pi}}}

\newcommand{\bxi}{\boldsymbol{\xi}}

\title{Distributionally Robust Multi-Output Regression Ranking}

\author[1]{Shahabeddin Sotudian}
\author[1]{Ruidi Chen}
\author[1,2]{Ioannis Ch. Paschalidis}
\affil[1]{Department of Electrical and Computer Engineering, Division of Systems Engineering}
\affil[2]{Department of Biomedical Engineering, and Faculty of Computing \& Data Sciences, \protect \\ Boston University, Boston, MA}

\begin{document}
\date{}
\maketitle

\begin{abstract}
Despite their empirical success, most existing listwise learning-to-rank (LTR) models are not built to be robust to errors in labeling or annotation, distributional data shift, or adversarial data perturbations. To fill this gap, we introduce a new listwise LTR model called Distributionally Robust Multi-output Regression Ranking (DRMRR). Different from existing methods, the scoring function of DRMRR was designed as a multivariate mapping from a feature vector to a vector of deviation scores, which captures local context information and cross-document interactions. DRMRR uses a Distributionally Robust Optimization (DRO) framework to minimize a multi-output loss function
under the most adverse distributions in the neighborhood of the empirical data distribution defined by a Wasserstein ball. We show that this is equivalent to a regularized regression problem with a matrix norm regularizer. Our experiments were conducted on two real-world applications, medical document retrieval, and drug response prediction, showing that DRMRR notably outperforms state-of-the-art LTR models. We also conducted a comprehensive analysis to assess the resilience of DRMRR against various types of noise: Gaussian noise, adversarial perturbations, and label poisoning. We show that DRMRR is not only able to achieve significantly better performance than other baselines, but it can maintain a relatively stable performance as more noise is added to the data. 
\end{abstract}

\section{Introduction}
There exist many real-world applications such as recommendation systems, document retrieval, machine translation, and computational biology where the correct ordering of instances is of equal or greater importance than minimizing regression or classification errors \cite{ru2021application}. \emph{Learning-to-rank (LTR)} refers to a group of algorithms that apply machine learning techniques to tackle these ranking problems. Generally speaking, LTR methods learn a scoring function that maps an instance-query feature vector to a relevance score (i.e., multi-level rating/label) that is then used to rank instances for a given query. Ideally, the resulting ranked list should maximize a ranking metric \cite{qin2010letor,sotudian2021improved,bruch2021alternative}.
We considered two medical applications of LTR, namely \emph{medical document retrieval} and \emph{drug response prediction}. Healthcare applications commonly face various challenges including $(i)$ susceptibilities in data collection due to instrument and environmental noise or data entry errors; $(ii)$ ambiguous or improper data annotation; $(iii)$ lack of large-scale data for training and testing of algorithms; $(iv)$ imbalanced data sets; $(v)$ missing data; $(vi)$ divergence of training and testing data distributions (e.g., data is recorded by different hospitals using different procedures); and more importantly, $(vii)$ the threat of adversarial attacks \cite{papangelou2018toward,finlayson2019adversarial,qayyum2020secure}. Consequently, robustness is critical for the wider adoption and deployment of algorithms into healthcare systems \cite{qayyum2020secure}.

In this work, without loss of generality, we take document retrieval as an example to explain the concepts and formulations. The main goal of document retrieval is to rank a set of documents by their relevance to a query. A slightly different example in computational biology is drug response prediction. Prescribing the right therapeutic option for each cancer patient is an intricate task since the efficacy of cancer medications varies among patients. Nevertheless, the biological differences among patients' cancers can be used to design genomic predictors of drug responses from large panels of cancer cell lines~\cite{sotudian2021machine}. 
In drug response prediction, large-scale screenings of cancer cell lines against libraries of pharmacological compounds are used to predict precise and individualized medications. 

Existing LTR approaches fall into three categories, namely pointwise, pairwise, and listwise~\cite{liu2011learning}. The pointwise approach formulates ranking as classification or regression techniques -- most early LTR algorithms such as linear regression ranking~\cite{liu2011learning} or RankNet~\cite{burges2005learning} take a very similar approach. In the pairwise approach, a classification method is employed to classify the preference order within document pairs. Representative pairwise ranking algorithms include RankBoost~\cite{freund2003efficient}, RankNet~\cite{burges2005learning}, and ordinal regression~\cite{liu2011learning}. Both approaches are misaligned with the ranking utilities such as Normalized Discounted Cumulative Gain (NDCG) and do not straightforwardly model the ranking problem. The listwise models can overcome this drawback by taking the entire list of retrieved documents for a query as instances and train a ranking function through the minimization of a listwise loss function. Experimental results show that the listwise approaches generally
outperform the pointwise and pairwise algorithms~\cite{cao2007learning}.
The literature offers a variety of approaches from deriving a smooth approximation to ranking utilities (e.g., ApproxNDCG~\cite{qin2010general} and SoftRank~\cite{taylor2008softrank}), to constructing differentiable surrogate loss functions (e.g., ListMLE~\cite{xia2008listwise}, LamdaMART~\cite{wu2010adapting}, and ListNet~\cite{cao2007learning}).

Most existing studies on LTR achieve impressive performance but often neglect the importance of \textbf{\emph{robustness}}~\cite{liu2011learning}. Systematic noise can become part of a data set in many ways and deceive LTR models to rank an item at an incorrect position with high confidence. While Empirical Risk Minimization (ERM) has been effective to optimize loss, ERM often does not yield models that are robust to adversarially crafted samples~\cite{biggio2013evasion}. \emph{Distributionally Robust Optimization (DRO)} is a modeling paradigm for data-driven decision-making under uncertainty. It has been successful in handling problems with corrupted training data through hedging against the most adverse distribution within a Wasserstein ball~\cite{chen2020distributionally}. Recently, DRO has been an active area of research owing to its robustness to adversarial examples, rigorous out-of-sample and asymptotic consistency guarantees, and remarkable empirical performance~\cite{sinha2017certifying}.

In the present work, we seek to induce robustness into LTR problems through using the DRO framework. Equipped with this perspective, we make the following contributions. Unlike other LTR frameworks, our algorithm approaches listwise ranking in a novel way and employs ranking metrics (i.e., NDCG) in its \emph{output}. In particular, we use the notion of \emph{position deviation} to define a vector of relevance scores instead of a scalar. We then adopt the DRO framework to minimize a worst-case expected multi-output loss function over a probabilistic ambiguity set that is defined by the Wasserstein metric. To the best of our knowledge, ours is the first study that utilizes a multi-output Wasserstein DRO framework to robustify LTR problems. We present an equivalent convex reformulation of the DRO problem, which is shown to be tighter than earlier work~\cite{chen2020robustified}. In experiments, our approach yields state-of-the-art results in two challenging applications of LTR, namely medical document retrieval and drug response prediction. More importantly, we evaluated our model to verify its robustness against various types of attacks including adversarial attacks and label attacks, showing that our model maintains a consistently good performance under various attack scenarios.

\textbf{Notational conventions:}  
We use boldfaced lowercase letters to
denote vectors, ordinary lowercase letters to denote scalars, boldfaced
uppercase letters to denote matrices, and calligraphic capital letters
to denote sets. All vectors are column vectors. For space
saving reasons, we write $\bx=(x_1, \ldots, x_{\text{dim}(\bx)})$ to
denote the column vector $\bx$, where $\text{dim}(\bx)$ is the dimension
of $\bx$. We use prime to denote the transpose, $\llbracket N \rrbracket$ for the set $\{1,\ldots,N\}$ for any integer $N$, $\|\cdot\|_p$
for the $\ell_p$ norm with $p \ge 1$, and $\bI_{K}$ for the $K$-dimensional identity matrix. For a matrix $\bA \in \mbb{R}^{m \times n}$, we use $\|\bA\|_p$ to denote its induced $\ell_p$ norm that is defined as $\|\bA\|_p \triangleq \sup_{\bx \neq \mathbf{0}}\|\bA\bx\|_p/\|\bx\|_p$.

\section{Preliminaries}

\subsection{Learning-to-Rank}
In a ranking problem, the data consists of a set of triples (query, document, relevance score). A feature vector is used to represent a query-document pair. The relevance score indicates the degree of relevance of this document to its corresponding query. 
Given a ranking data set ${\{(\bX^{q},\btheta^{q})\}}_{q=1}^{T}$, $q \in \llbracket T \rrbracket$ indexes a query, and $\bX^{q}$ and $\btheta^{q}$ represent  the list of retrieved documents and corresponding relevance scores, respectively. The ${q}$-th query contains ${n_{q}}$ documents and  $\bX^{q}\in \mathbb{R}^{{n_{q}} \times p}$
has rows 
$( \bx_{1}^{q},\cdots, \bx_{n_{q}}^{q} )$, each of which is a $p$-dimensional document feature vector. The vector $ \btheta^{q} =( \theta_{1}^{q},\cdots, \theta_{n_{q}}^{q} ) \in \mathbb{R}_{+}^{n_{q}} $ contains the corresponding ground-truth relevance scores, where a higher $\theta_{d}^{q}\in \mathbb{R}$ indicates that the document with features $\bx_{d}^{q}$ is more relevant.
In the learning-to-rank framework, denoting by $\bx$ and $\theta$ the random variables that represent the document feature vector and relevance score, respectively, the goal is to learn a scoring function $f$ that best predicts the relevance score: 
\begin{equation} \label{Prel:eq0}
   \min_{f}\mathcal{L}(f) \triangleq \mbb{E}^{\mbb{P}^*} [\ell(\theta, f(\bx))],
\end{equation}
where $\ell: \mbb{R} \times \mbb{R} \rightarrow \mbb{R}$ is a loss function, $f: \mbb{R}^p \rightarrow \mbb{R}$ predicts the relevance score of each document, and $\mbb{P}^*$ is the underlying true probability distribution of $(\bx, \theta)$. Given that $\mbb{P}^*$ is unknown, most existing LTR algorithms solve (\ref{Prel:eq0}) through estimating the expected loss by its empirical substitute (\ref{Prel:eq1}): 
\begin{equation} \label{Prel:eq1}
    \hat{\mathcal{L}}(f) \triangleq \frac{1}{\sum_{q} n_q} \sum_{q=1}^T \sum_{d=1}^{n_q} \ell\big(\theta_d^q, f(\bx^q_d)\big).
\end{equation}
For a test query $\bX^t \in \mbb{R}^{n_t \times p}$ consisting of $n_t$ documents, the final predicted ranking list $\hat{\bpi}$ is simply obtained by ranking the rows in $\bX^t$ based on their inferred ranking scores $\hat{\btheta}^t = (f(\bx_1^t), \ldots, f(\bx_{n_t}^t))$.
 Eq.~(\ref{Prel:eq1}) is restrictive in the sense that $(i)$ it does not take into account the inter-dependency of scores between documents, and $(ii)$ the empirical estimate is very sensitive to data perturbations. 

\subsection{Distributionally Robust Optimization}
Distributionally Robust optimization (DRO) hedges against a set of probability distributions instead of just the empirical distribution. DRO minimizes a worst-case loss over a probabilistic ambiguity set:
\begin{equation*}
\min_{f} \max_{\mathbb{Q} \in \Omega}  \mbb{E}^{\mathbb{Q}} \left[ \ell(\theta, f(\bx))\right],
\end{equation*}
where the ambiguity set $\Omega$ can be defined through moment constraints~\cite{wiesemann2014distributionally}, or as a ball of
distributions using some probabilistic distance function such as the Wasserstein distance~\cite{blanchet2019quantifying,shafieezadeh2019regularization}. The Wasserstein DRO model has been extensively studied in the machine learning community; see, for example, \cite{blanchet2019multivariate, chen2018robust} for robustified regression models, \cite{sinha2017certifying} for adversarial training in neural networks,
and \cite{shafieezadeh2015distributionally} for distributionally robust logistic regression. These works and  \cite{gao2017wasserstein,chen2020distributionally} provided a comprehensive analysis of the Wasserstein-based distributionally robust statistical learning framework.

\section{Problem Formulation}
Next, we introduce our DRO formulation of the LTR problem. Different from the existing works where a univariate relevance score $\theta_d^q \in \mbb{R}$ is used for each document $\bx_d^q \in \mbb{R}^p$, we define a Ground Truth Deviation vector $\btheta_d^q \in \mbb{R}^K$ to characterize different levels of importance for the document $\bx_d^q$ in the $q$-th query. Here, $K$ is a constant to be defined later (end of Sec.~\ref{sec:GTD}). We derive an equivalent reformulation of the DRO problem in Sec.~\ref{sec:DRMRR}.

\subsection{Ground Truth Deviation} \label{sec:GTD}
As a popular evaluation criterion in information retrieval, Normalized Discounted Cumulative Gain (NDCG) can deal with cases that have more than two degrees of relevancy for documents \cite{ravikumar2011ndcg}. Let $D(s)=\frac{1}{\log(1+s)}$ be a discount function, $G(s) = s$, a monotone increasing gain function, and $ \scrZ_n = \lbrace (\mathbf{x}_1, y_1), ..., (\mathbf{x}_n, y_n)  \rbrace$ a set of documents ordered according to their ground-truth rank, with $\mathbf{x}_i$ and $y_i$ being a document feature vector and a relevance score, respectively.
Assume $\tilde{\scrZ}_n$ is a (predicted) ranked list for $\scrZ_n$; then
the Discounted Cumulative Gain (DCG) of $\tilde{\scrZ}_n$ is defined as $\Phi(\tilde{\scrZ}_n) = \sum_{r=1}^n G(y_{\pi_r})D(r)$, where $\pi_r$ is the index of the document ranked at position $r$ of $\tilde{\scrZ}_n$. The reason for introducing the discount function is that the user cares less about documents  ranked lower~\cite{wang2013theoretical}. NDCG normalizes DCG by the Ideal DCG (IDCG), $\Phi^{I}({\scrZ}_n) $, which is the DCG score of the ideal ranking result \cite{valizadegan2009learning} and can be computed by $\Phi^{N}(\tilde{\scrZ}_n) = \frac{\Phi(\tilde{\scrZ}_n)}{\Phi^{I}({\scrZ}_n)}\in [0,1]$.

Considering the ${q}$-th query  $(\bX^{q},\by^{q})$ that contains ${n_{q}}$ documents, we define Ground Truth Deviation (GTD) vector for document $d$ as follows:
\begin{equation}\label{PF:eq7}
\btheta_{d}^{q} = \xi_{I}({\bxi}_{D}\circ {\bxi}_{\Phi}),
\end{equation}
where $\circ$ is the Hadamard product (a.k.a. the element-wise product). The vector $\btheta_{d}^{q}$ is comprised of the following three components: 

\paragraph{\textbf{NDCG deviation score (${\bxi}_{\Phi}$).}} To compute this vector, first, the elements of $ {\by}^{q} =( y_{1}^{q},\cdots, y_{n_{q}}^{q} )$ are sorted in descending order of their individual relevance scores, and the document feature vectors $ \bX^{q} =( \bx_{1}^{q},\cdots, \bx_{n_{q}}^{q} )$ are also sorted correspondingly. We denote them by $\Bar{\by}^{q}$ and $\Bar{\bX}^{q}$, respectively. The NDCG score for $\Bar{\bX}^{q}$ is equal to 1. If we switch two documents in $\Bar{\bX}^{q}$, the NDCG will decrease or in some cases may stay the same (i.e., if their relevance scores are equal). For document $d$ in query $q$, we define NDCG deviation score vector as ${\bxi}_{\Phi}=\left[ {\lambda_{d1}}, {\lambda_{d2}}, ..., {\lambda_{d{n_{q}}}}   \right ]$ where ${\lambda_{di}}$ is the NDCG score of $\Bar{\bX}^{q}$ when we switch the \textbf{\emph{position}} of document $d$ with the \textbf{\emph{document}} that is in $i$-th position of $\Bar{\bX}^{q}$ and can be formulated as follows:
\begin{equation*}
 {\lambda_{di}} = 1 + \frac{ \frac{y_d - y_{\pi_i}}{\log(1+i)}  + \frac{y_{\pi_i} - y_d}{\log(1+\pi_{d}^{-1})}}{\Phi^{I}}.
\end{equation*} 
Here, $\pi_{d}^{-1}$ is the position of the document $d$ in $\Bar{\bX}^{q}$, $\pi_i$ is the index of the document ranked at the $i$-th position of $\Bar{\bX}^{q}$, and $\Phi^{I}$ is the IDCG. The details about the derivation can be found in the supplementary materials. We can perceive the $i$-th element of the GTD vector as a score that indicates the degree of congruence between a document and the $i$-th rank.

\begin{figure}[!tb]
	\centering
	\subfloat[$\ {\bxi}_{D}$]{\includegraphics[width=0.23\textwidth]{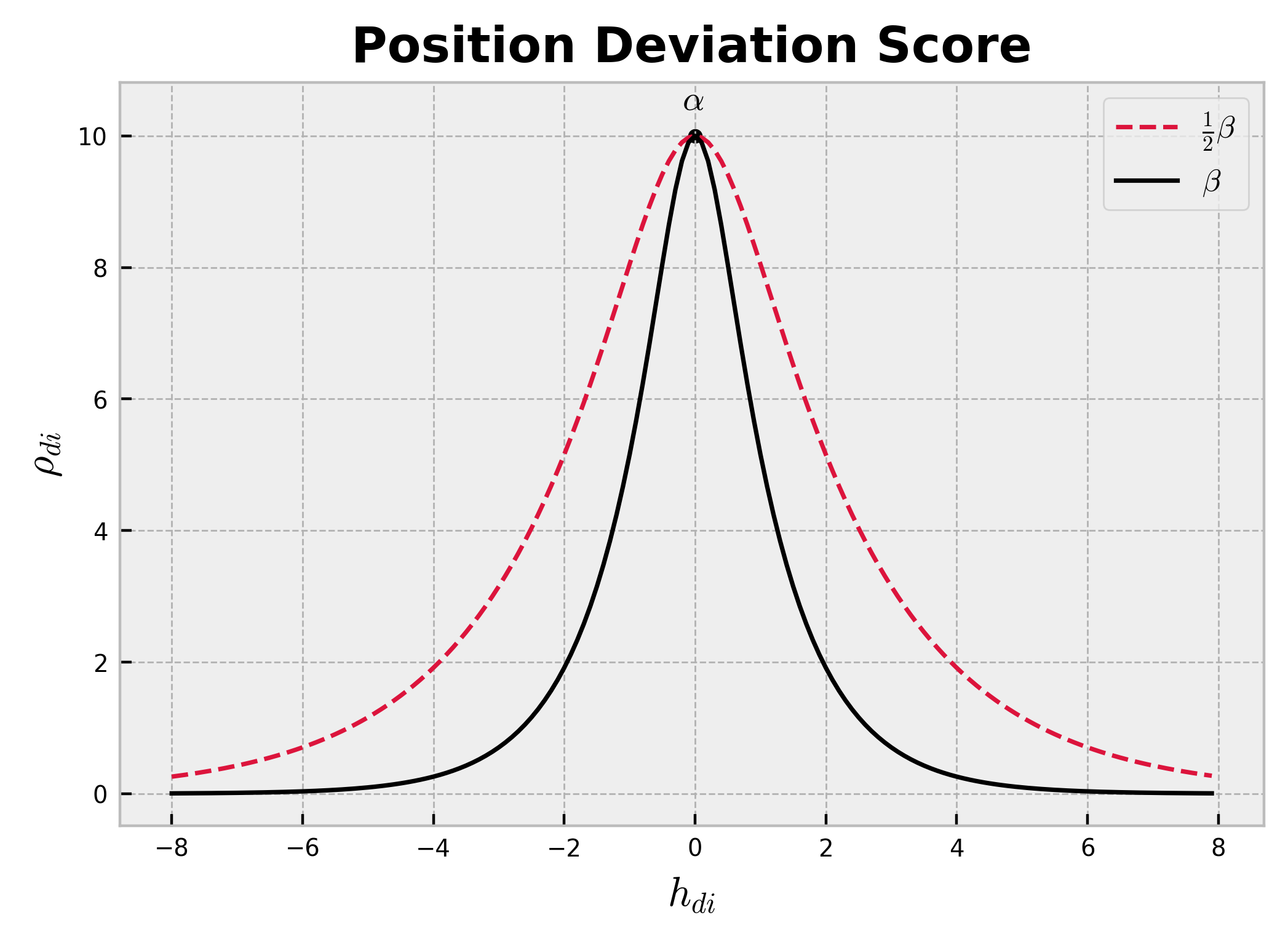}}\hfill
	\subfloat[$\ {\xi_{I}}$]{\includegraphics[width=0.23\textwidth]{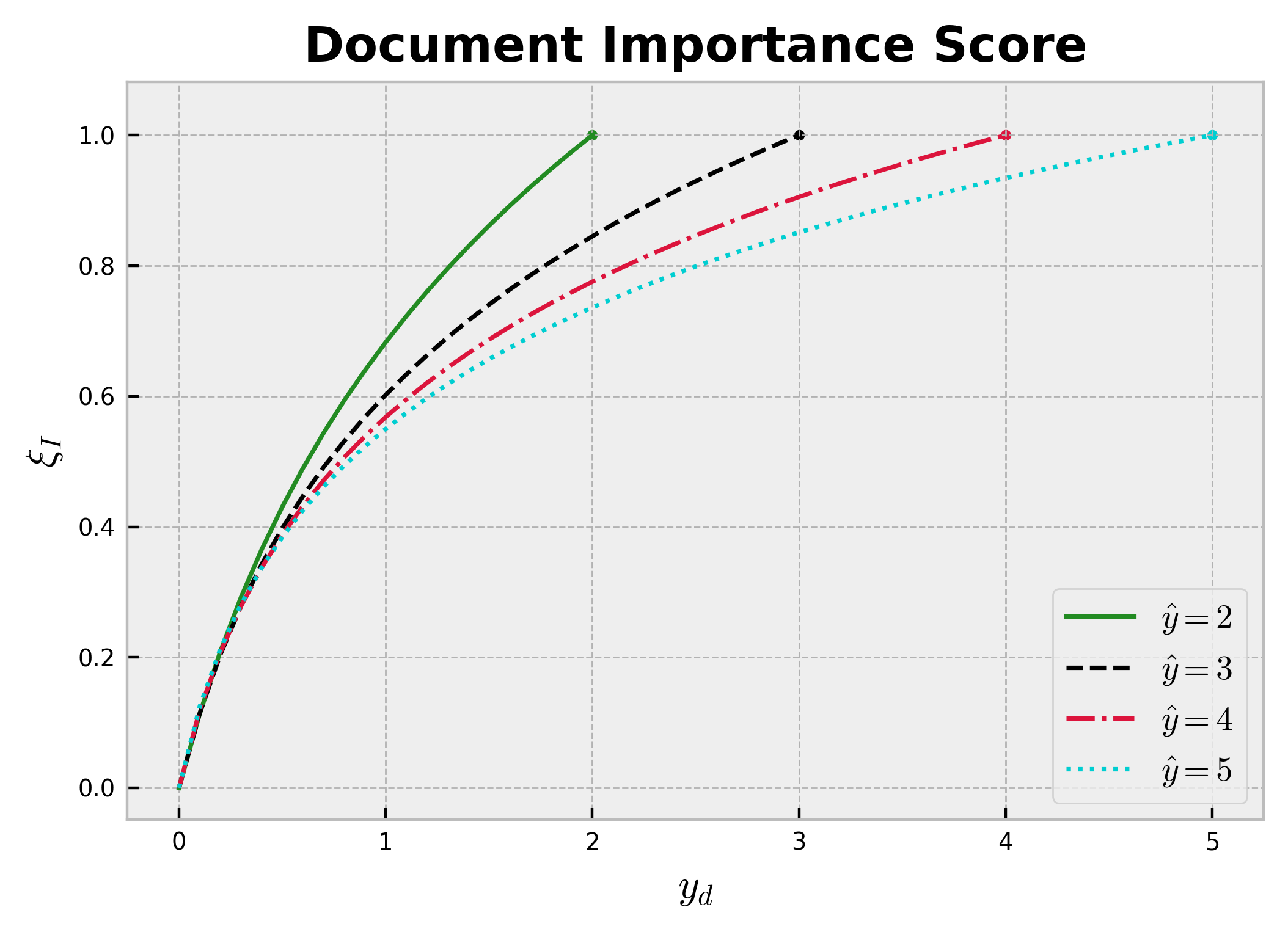}}
	\caption{GTD graphs: (a) Position deviation score where $\alpha=10$ and $\beta=2$. (b) Document importance score for various maximum possible relevance scores.}\label{fig:GTD}
\end{figure}

\paragraph{\textbf{Position deviation score (${\bxi}_{D}$).} } This vector is defined to further push the relevant documents to the top of the ranking list, and penalize documents based on their position in the ranking list. Position deviation score works in conjunction with ${\bxi}_{\Phi}$. We defined it as ${\bxi}_{D}=\left[ {\rho_{d1}}, {\rho_{d2}}, ..., {\rho_{d{n_{q}}}}   \right ]$ where ${\rho_{di}}$ can be calculated by
\begin{equation*}
\rho_{di} = \frac{\alpha}{\sqrt{\left|cosh\left(min \left(\beta h_{di},\frac{\beta}{2}h_{di}\right)\right)\right|}},
\end{equation*} 
where $h_{di}= \pi_{d}^{-1} - i$. As can be seen in Figure \ref{fig:GTD}, $\alpha$ specifies the GTD's maximum score and $\beta$ regulates the magnitude of the penalty for a position deviation. Here, we use the red dashed curve for positive deviations (i.e., when a document ranked higher than its optimal position) and the black curve for negative deviations. This would induce our model to tolerate a positive deviation more than a negative one. Consequently, the model pushes the relevant documents to the top of the ranking list.

\paragraph{\textbf{Document importance score (${\xi_{I}}$).}} It is defined to place greater emphasis on highly relevant documents and can be computed as follows:
\begin{equation*}
 {\xi_{I}} = \frac{\log(y_d \cdot \hat{y}+1)}{\log(\hat{y}^{2}+1)},
\end{equation*} 
where $\hat{y}$ is the maximum possible value for relevance scores. Figure \ref{fig:GTD} presents $\xi_{I}$ for different values of $\hat{y}$.\par

Ultimately, instead of a relevance score for each document, we have a GTD vector. The GTD vector characterizes different levels of importance for a document in a query where the first element is the first level of importance, the second element is the second level of importance, and so on. Since each query may have different number of documents, we just consider the first $K$ elements of ${\bxi}_{\Phi}$ and ${\bxi}_{D}$ in our model, corresponding to $K$ levels of importance.  In this way, all GTD vectors are the same length. We prefer to use a low value for $K$ since it forces the model to focus on the most relevant documents. In case a large $K$ needs to be used and $K>n_{q}$, we can simply repeat the last element of ${\bxi}_{\Phi}$ and ${\bxi}_{D}$ to pad our ${\bxi}_{\Phi}$ vector.

\subsection{Distributionally Robust Multi-Output Regression} \label{sec:DRMRR}
Consider a setting where there are $K$ levels of importance with features and importance scores distributed according to $\bx \in \mbb{R}^p$ and  $\btheta \in \mbb{R}^K$, respectively. 
We restrict our attention to linear function classes by assuming $\mathbf{f}(\bx) = \bB'\bx$ where $\bB \in \mbb{R}^{p \times K}$. The matrix $\bB$ characterizes the dependency structure of the different levels of importance. Nonlinearity can be introduced by applying a transformation (e.g., kernel function) on the feature $\bx$. The Distributionally Robust Multi-output Regression Ranking (DRMRR) formulation minimizes the worst-case expected loss as follows:
\begin{equation}\label{PF:eq1}
 \inf_{\bB} \sup_{\mathbb{Q} \in \Omega}  \mathbb{E}^{\mathbb{Q}} \big[ \ell (\btheta - \bB'\bx)\big],
\end{equation}
where $\ell: \mathbb{R}^{K} \to \mathbb{R}$ is a Lipschitz continuous loss function on the metric spaces $(\mathcal{D}, \lVert \cdot \rVert_{r} )$ and $(\mathcal{C}, \lvert \cdot \rvert )$, where $\mathcal{D}$, $\mathcal{C}$ are the domain and co-domain of $\ell (\cdot)$, respectively. $\mathbb{Q} \in \Omega \triangleq \lbrace \mathbb{Q} \in \scrP(\scrS) : \mathit{W}_{1}(\mathbb{Q},\hat{\mathbb{P}}_{N} )  \leq \varepsilon   \rbrace$ is the probability distribution of $(\bx,\btheta)$, where $\scrP(\scrS)$ is the space of all probability distributions supported on $\scrS$ and $\scrS$ is the uncertainty set of $(\bx, \btheta)$, $\varepsilon$ is a positive constant (i.e., Wasserstein ball radius), $\hat{\mathbb{P}}_{N}$ is the empirical distribution that assigns an equal probability to all $N$ training samples, with $N = \sum_{e=1}^T n_{e} $, where $T$ is the number of queries, and $\mathit{W}_{1}(\mathbb{Q},\hat{\mathbb{P}}_{N} )$ is the order-1 Wasserstein distance between $\mathbb{Q}$ and $\hat{\mathbb{P}}_{N}$ defined as follows:
\begin{equation*}
 \mathit{W}_{1}(\mathbb{Q},\hat{\mathbb{P}}_{N} ) \triangleq 
 \min_{\Pi \in P(\scrS \times \scrS)} \Bigl\lbrace \int_{\scrS \times \scrS}  
 \delta ( \bz_{1} -  \bz_{2}) 
 \Pi (d\bz_{1}, d\bz_{2})
 \Bigr\rbrace.
\end{equation*}
In $\mathit{W}_{1}(\mathbb{Q},\hat{\mathbb{P}}_{N} )$, $\delta ( \bz_{1} -  \bz_{2}) \triangleq \lVert \bz_{1} -  \bz_{2} \rVert_{r}$ with $\bz_i = (\bx_i,\btheta_i), i=1,2$, drawn from $\mbb{Q}$ and $\hat{\mbb{P}}_N$, respectively, and $\Pi$ specifies the joint distribution of $\bz_{1}$ and $\bz_{2}$ with marginals $\mathbb{Q}$ and $\hat{\mathbb{P}}_{N}$. Note that the same norm is used to define the Wasserstein metric and the domain space of $\ell (\cdot)$. In the following theorem we propose an equivalent reformulation of (\ref{PF:eq1}) by using duality for the inner maximization problem. 
\begin{theorem}
	\label{Theorem:CP_DRO_reformulation}
	Suppose our dataset consists of $T$ queries $\big\{(\bX^q, \bTheta^q)\big\}_{q=1}^{T}$ and each query $q$ contains $n_{q}$ documents, $q \in \llbracket T \rrbracket$, where $ \bX^q \in \mathbb{R}^{n_{q} \times p}$ is the document feature matrix with rows $\bx_d^q \in \mbb{R}^p$, $d \in \llbracket n_{q} \rrbracket$, and $\bTheta^q \in \mbb{R}^{n_{q} \times K}$ is the GTD matrix with rows $ \btheta_d^q \in \mathbb{R}^{K}$. Define a loss function $ \ell(\cdot) \triangleq \|\cdot\|_r$. If the Wasserstein metric is induced by $\lVert \cdot \rVert_{r}$,  the DRMRR problem (\ref{PF:eq1}) can be equivalently reformulated as:
	\begin{equation}\label{PF:eq4}
	\inf_{\bB} \  \frac{1}{\sum_{e=1}^T n_{e}} \sum_{q=1}^{T} \sum_{d=1}^{n_{q}} \|\btheta_d^q-\bB'\bx_d^q\|_r 
	+ \varepsilon \lVert \tilde{\bB}' \rVert_{s},
	\end{equation}
	where $r,s\geq1$; $1/r+1/s=1$; $\tilde{\bB} = (-\bB',\bI_{K} )$.
\end{theorem}
The proof can be found in the supplementary materials. Thm.~\ref{Theorem:CP_DRO_reformulation} establishes a connection between distributional robustness and regularization, which has also been studied by, e.g.,  \cite{shafieezadeh2015distributionally, shafieezadeh2019regularization,gao2017wasserstein}. However, most of the existing studies focused on a univariate output. By contrast, our work adapts the DRO framework to a multi-output setting, which is more suitable for the ranking problem. Recently, \cite{chen2020robustified} studied a multi-output regression problem under the Wasserstein DRO framework. However, our results in Theorem \ref{Theorem:CP_DRO_reformulation} present a tighter reformulation than theirs. In the case where the Wasserstein metric is induced by the $\ell_2$ norm ($r=2$), Eq.~(\ref{PF:eq4}) yields a regularizer which is the spectral norm (largest singular value) of $\tilde{\bB}'$, while \cite{chen2020robustified} derived a regularizer in the Frobenius norm which is looser. 

\subsection{Score calculation}
Suppose we are given a test query $ \bX^{t} =( \bx_{1}^{t},\cdots, \bx_{n_{t}}^{t} ) \in \mathbb{R}^{{n_{t}} \times p} $; we can estimate the GTD matrix as $\hat{\bTheta}^{t} = \left(\bB^{'}\bx_1^{t},\dots, \bB^{'}\bx_{n_t}^{t}\right)\in \mathbb{R}^{n_{t}\times K}$. In matrix $\hat{\bTheta}^{t}$, columns correspond different ranks and rows refers to different documents. Algorithm \ref{alg1} demonstrates the procedure of ranking using the output of the DRMRR algorithm where $R_{K}(j)$ is the remainder of dividing $j$ by $K$.

\begin{algorithm}[!bt]
\caption{\textbf{Scoring procedure for DRMRR}}\label{alg1}
\begin{algorithmic}[] 
\STATE \textbf{Input}: $\hat{\bTheta}^{t}$
\STATE \textbf{Output}: Sorted list
\STATE \textbf{Let} $\delta = 1$;
\FOR{$j=1$ \TO ${n_t}$}
    \STATE Find the maximum of $\delta$-th column of $\hat{\bTheta}^{t}$;
    \STATE Assign the corresponding row/document to rank $j$;
    \STATE Remove the corresponding row/document;
    \IF {$R_{K}(j) = 0$}
        \STATE $\delta = 1$;
    \ELSE
        \STATE  $\delta = \delta + 1$;
    \ENDIF 
\ENDFOR 
\end{algorithmic}
\end{algorithm}
\section{Experimental Results}  \label{sec:exp}

\subsection{Experiment setup}\label{sec:setup}
\paragraph{\textbf{Data sets:}} We conducted experiments on two publicly available benchmark datasets: OHSUMED and Drug Response Prediction (DRP). As a subset of the MEDLINE database (a database on medical publications), the OHSUMED corpus \cite{hersh1994ohsumed} consists of about 0.3 million records from 270 medical journals from 1987 to 1991. A query set with 106 queries on the OHSUMED corpus has been extensively used in previous works, in which each query is represented by 45 features \cite{qin2010letor}. There are in total 16,140 query-document pairs with relevance judgments. LETOR \cite{qin2010letor} defined three ratings 0, 1, 2, corresponding to “irrelevant,” “partially relevant,” and “definitely relevant,” respectively.
In addition to OHSUMED, we trained and evaluated our method using the cell line data and drug sensitivity data from the {\em Cancer Cell Line Encyclopedia
(CCLE)}~\cite{noauthor_broad_nodate} and the {\em Cancer Therapeutics Response Portal (CTRP
v2)}~\cite{noauthor_cancer_nodate}. A total of 332 cell lines and 50 drug responses were used. The ``Act Area'' (the area above the fitted dose-response curve) was used to quantify drug sensitivity where lower response value indicates higher drug sensitivity. After several pre-processing steps, cell lines are represented by 251 numeric features (i.e., genes) and drug sensitivities are labeled with graded relevance from 0 to 2 (i.e., ``insensitive,'' ``sensitive,'' and ``highly sensitive,'' respectively) with larger labels indicating a higher sensitivity. Further details of the pre-processing steps can be found in the supplement.


\paragraph{\textbf{Evaluation metrics:}} We evaluated model performance using two metrics:  NDCG@k and AP@k. NDCG@k is the top-k version of NDCG, where the discount function is $D(s) = 0$ for $s > k$. Precision at position k (P@k) is the fraction of relevant documents in the top-k.  Suppose we have binary relevance for the documents in a $q$-query; we define P@k as $P@k= \frac{1}{k} \sum_{j=1}^{k} \mathbbm{1}(y_{\pi_j}=1)$ where $\mathbbm{1}(\cdot)$  is the indicator function. We define Average Precision at position k as $AP@k = \frac{1}{m} \sum_{j=1}^{k} P@k\times \mathbbm{1}(y_{\pi_j}=1)$, where $m$ is the total number of relevant documents in the top-k of the ranking list. AP is a highly localized performance measure and captures the quality of rankings for applications where only the first few results matter. The main difference between AP and NDCG is that NDCG differentiates between “partially relevant” and “definitely relevant” documents while AP treats them equally. Given a set of testing queries and a performance metric, we are interested in the mean metric which is simply the mean of the performance metric for all queries. From now on, we use NDCG@k and AP@k to denote mean NDCG@k and mean AP@k, respectively.

\paragraph{\textbf{Competing Methods:}} Although the list of published LTR algorithms is endless, LamdaMART\textsubscript{MAP} \cite{wu2010adapting},  LamdaMART\textsubscript{NDCG} \cite{wu2010adapting}, and XE-MART\textsubscript{NDCG} \cite{bruch2021alternative} have been demonstrated repeatedly to outperfrom other algorithms including RankNet \cite{burges2005learning}, Coordinate Ascent \cite{metzler2007linear}, ListNet \cite{cao2007learning}, Random Forests \cite{breiman2001random}, BoltzRank \cite{volkovs2009boltzrank}, ListMLE \cite{xia2008listwise}, Position-Aware ListMLE \cite{lan2014position},  RankBoost \cite{freund2003efficient}, AdaRank \cite{xu2007adarank}, SoftRank \cite{taylor2008softrank}, ApproxNDCG \cite{qin2010general}, ApproxAP \cite{qin2010general}, and several direct optimization methods \cite{xu2008directly,metzler2005direct}. Here, we rely on prior research \cite{bruch2021alternative,wang2018lambdaloss} and do not include the weaker methods in our experiments. It is important to note that the author of XE-MART\textsubscript{NDCG} proposed this model as a \emph{robust alternative} to LamdaMART-based models. We used open-source python packages (i.e., LightGBM \cite{ke2017lightgbm,meng2016communication,zhang2017gpu} and XGBoost \cite{chen2016xgboost}) to implement the baseline models.

\paragraph{\textbf{Experimental settings and hyper-parameter optimization:}} In our experiments, we followed the standard supervised LTR framework \cite{liu2011learning}. Authors of LETOR~\cite{qin2010letor} partitioned the OHSUMED data set into five parts for five-fold cross-validation where three parts are used for training, one part for validation (i.e., tuning the hyperparameters of the learning algorithms), and the remaining part for evaluating the performance of the learned model. Similarly, we partitioned the drug response data set into five folds and conducted five-fold cross-validation to train, validate and evaluate the ranking algorithms. In all experiments, the average on the test set over the 5 folds was reported. Algorithm parameters were tuned on the validation sets. We optimized the algorithm parameters to maximize NDCG@5 and NDCG@10. The details of the parameter-tuning procedure and the optimal parameters for each algorithm can be found in the supplement.

\subsection{Overall Comparison}\label{Sec_OveralCompar}
We compared the performance of the proposed method on OHSUMED and Drug Response Prediction (DRP) data sets with baseline methods introduced in
Sec.~\ref{sec:setup}. The results are in Table \ref{table:tab1}. The values inside the parentheses denote the Standard Deviation (SD) of the corresponding metrics. Bold numbers indicate the best performance among all methods for each metric. DRMRR consistently outperforms all baseline methods across all  metrics.
In our experiment on OHSUMED data, LamdaMART\textsubscript{NDCG}  demonstrated a reasonably good overall performance and it is the second-best method. However, XE-MART\textsubscript{NDCG} was the second-best method in our experiment on the DRP data. The difference between the best and the second-best methods for the DRP data set is greater than what we obtained for OHSUMED. Due to the limited number of samples available and the specific structure of the DRP data, the performance of the baseline methods diminished significantly. On the other hand, DRMRR was able to maintain its high performance. To sum up, the proposed method is not only able to push the most relevant documents (or sensitive drugs) to the top of the ranking list, but it can put them in the right order.

\begin{table*}[!tb]
\centering
\resizebox{0.9\linewidth}{!}{
\begin{tabular}{clllll}
\thickhline
\multicolumn{1}{l}{}                & \textbf{Algorithms} & \textbf{NDCG@5} & \textbf{NDCG@10}  & \textbf{AP@5} & \textbf{AP@10} \\ \thickhline
\multirow{4}{*}{\textbf{OHSUMED}}   & LamdaMART\textsubscript{MAP}          & 45.18\% (5.07\%)         & 43.65\% (3.55\%)         & 67.94\% (7.26\%)        & 64.12\% (5.83\%)         \\
                                    & LamdaMART\textsubscript{NDCG}         & 46.17\% (5.91\%)         & 44.40\% (5.00\%)         & 68.53\% (8.40\%)        & 65.25\% (6.47\%)         \\
                                    & XE-MART\textsubscript{NDCG}           & 44.31\% (6.58\%)         & 44.79\% (5.65\%)         & 65.25\% (8.08\%)        & 62.41\% (6.76\%)         \\
                                    & DRMRR                                 & \textbf{47.79\% (6.58\%)}         & \textbf{45.36\% (4.84\%)}          & \textbf{70.84\% (7.35\%)}        & \textbf{65.31\% (7.35\%)}         \\ \thickhline
\multirow{4}{*}{\textbf{DRP}}       & LamdaMART\textsubscript{MAP}          & 58.11\% (1.90\%)         & 63.39\% (2.17\%)         & 83.27\% (1.57\%)        & 77.25\% (1.07\%)         \\
                                    & LamdaMART\textsubscript{NDCG}         & 58.73\% (2.54\%)         & 62.87\% (2.83\%)         & 83.07\% (1.99\%)        & 76.95\% (1.19\%)         \\
                                    & XE-MART\textsubscript{NDCG}           & 59.37\% (1.92\%)         & 63.51\% (2.18\%)         & 83.70\% (1.28\%)        & 77.43\% (1.34\%)         \\
                                    & DRMRR                                 & \textbf{68.40\% (1.74\%)}         & \textbf{71.27\% (1.78\%)}          & \textbf{85.03\% (1.10\%)}         & \textbf{81.03\% (1.00\%)}          \\ \thickhline
\end{tabular}}
\caption{Performance Comparison of Ranking Methods. }\label{table:tab1}
\end{table*}

\subsection{Robustness Comparison}
In this section, we empirically study the behavior of DRMRR in the presence of noise. While our overall performance analysis suggested that DRMRR should be the most ``well-behaved" of the four, that analysis was performed on the clean data. The robustness of a ranking model to noise is crucial in practice, especially in the healthcare domain. We put this hypothesis to test through four types of experiments. We conducted all experiments on the OHSUMED data set since it is a popular and standard LTR data set. We just reported AP@5 and NDCG@5, other metrics are presented in the supplementary material. In all experiments, the values are the average of 5 folds.

\begin{figure}[!tb]
	\centering
	\subfloat[AP@5]{\includegraphics[width=0.23\textwidth]{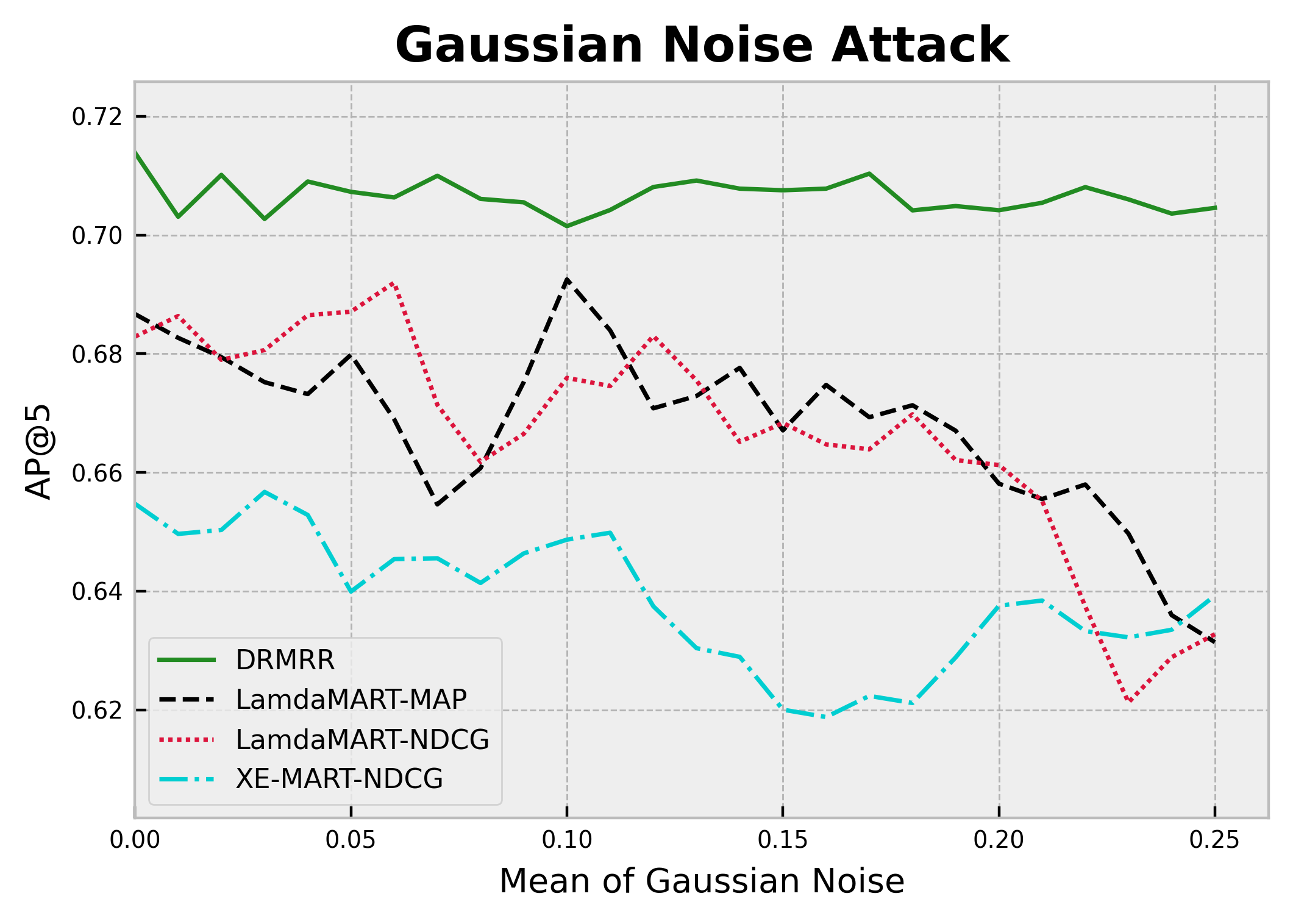}}\hfill
	\subfloat[NDCG@5]{\includegraphics[width=0.23\textwidth]{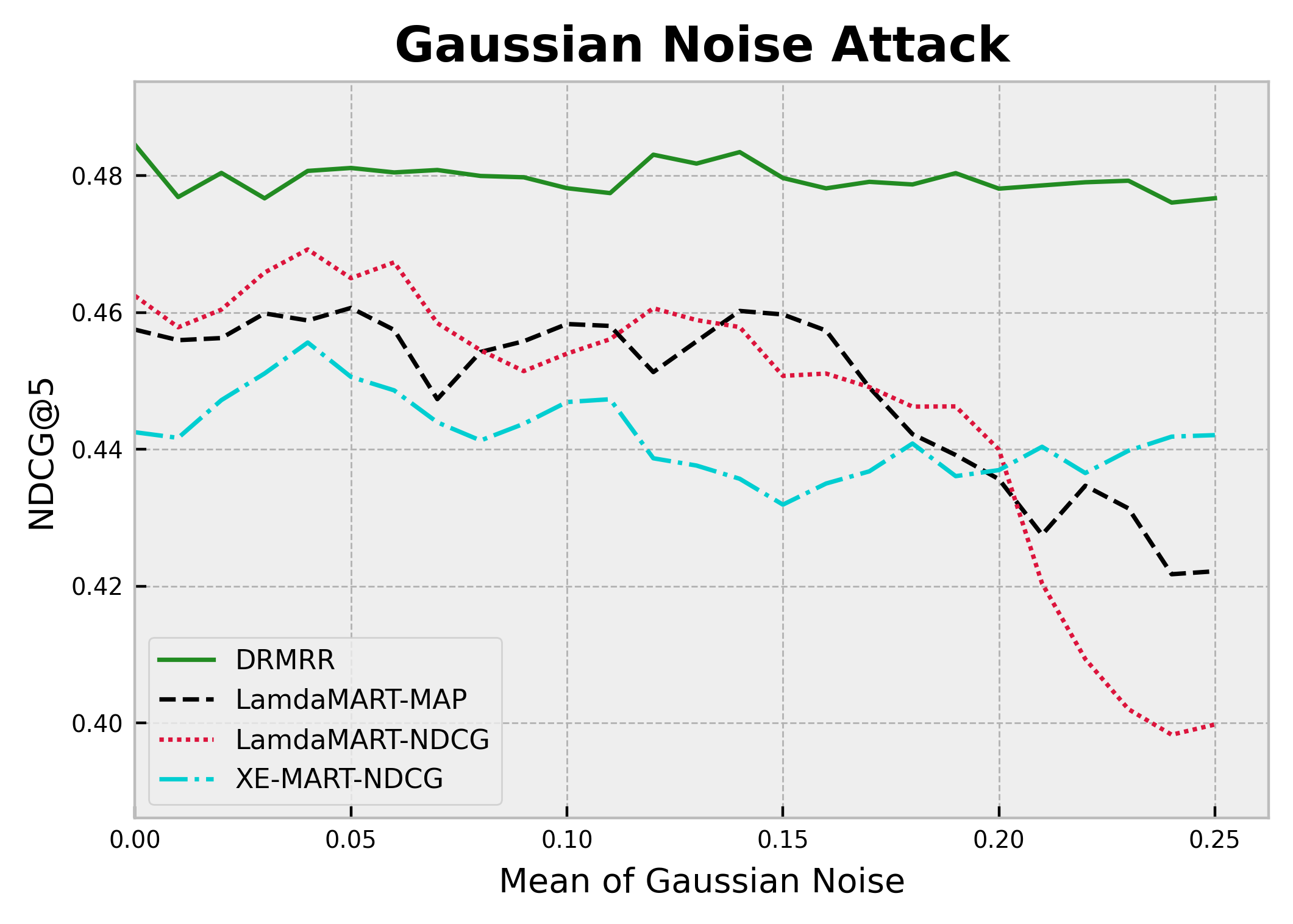}}	
	\caption{The impact of Gaussian noise on the performance of ranking models.}\label{fig:Robust_GN}
\end{figure}

\paragraph{\textbf{Gaussian Noise Attack:}}  We added Gaussian noise to the test documents to deliberately corrupt them; therefore, depreciating their predictability. Gaussian noise was added to 75\% of the test queries randomly. Experiments were conducted using various means and a fixed standard deviation of 0.001.
We used the perturbed test data to evaluate the trained models in Sec.~\ref{Sec_OveralCompar} (i.e., all algorithms were tested on the same perturbed test data). Fig.~\ref{fig:Robust_GN} demonstrates the performance of the algorithms on the perturbed test data. Two observations are in order: $(i)$ DRMRR outperformed the baseline models at different levels of noise; and $(ii)$ DRMRR demonstrated a relatively stable performance.

\paragraph{\textbf{Universal Adversarial Perturbation Attack:}} We built an adversarial model to introduce perturbations that break the neighborhood relationships by altering the input slightly. To that end, a pointwise linear regression ranking model was trained as the adversarial model on the clean training set. Then, 75\% of the test queries were perturbed using the coefficient of the adversarial model and the Fast Gradient Sign Method (FGSM) method: $\bar{\bar{\bx}}_{q}^{d} = {\bx}_{q}^{d} + \epsilon \cdot \text{sgn}(\nabla_{{\bx}_{q}^{d}}J({\bx}_{q}^{d},{y}_{q}^{d}))$ where $\bar{\bar{\bx}}_{q}^{d}$ is the perturbed feature vector, $\epsilon$ controls the magnitude of the perturbations, and $J$ is the cost function of the adversarial model~\cite{goodfellow2014explaining}.

\begin{figure}[!tb]
	\centering
	\subfloat[AP@5]{\includegraphics[width=0.23\textwidth]{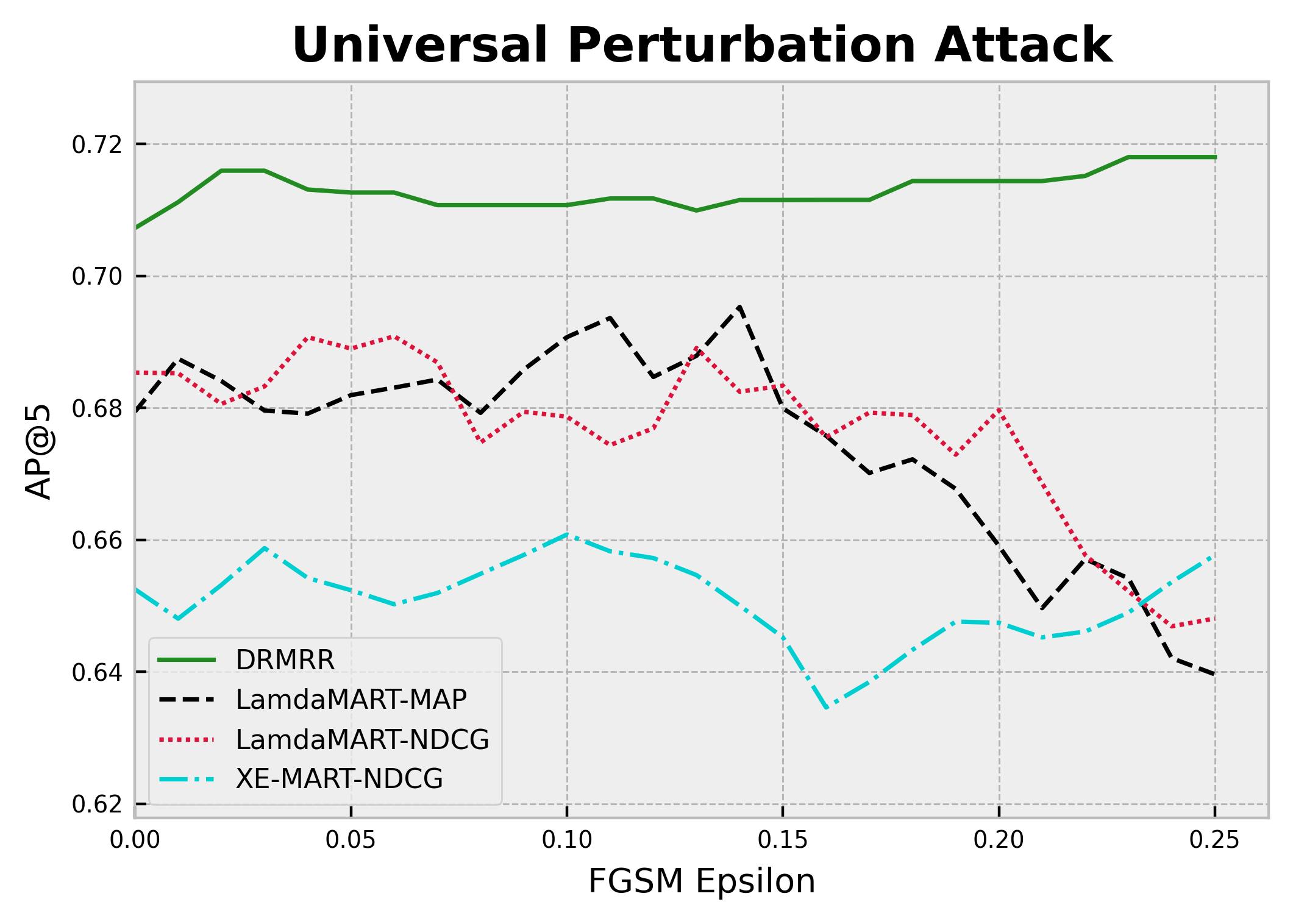}}\hfill
	\subfloat[NDCG@5]{\includegraphics[width=0.23\textwidth]{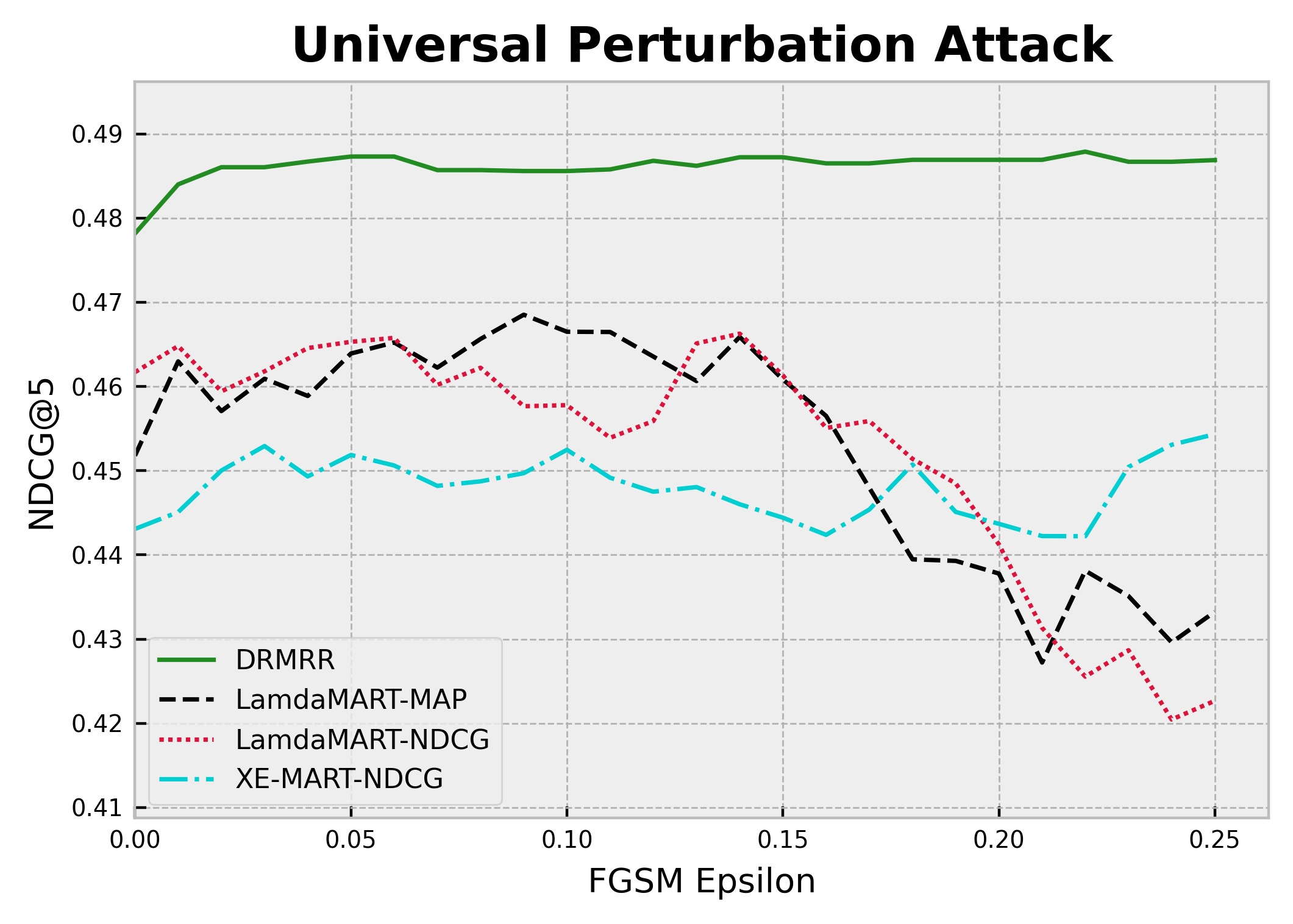}}
	\caption{The impact of universal adversarial perturbation on the performance of ranking models.}\label{fig:Robust_UAP}
\end{figure}

All algorithms that we trained in Sec.~\ref{Sec_OveralCompar} were evaluated on the same perturbed test data. In this case, the adversary had no knowledge of the ranking models; however, it was trained on the same training data. Fig.~\ref{fig:Robust_UAP} shows the performance of the algorithms on the perturbed test data. As we increase the level of perturbations (i.e., $\epsilon$), we can see that DRMRR is less sensitive to adversarial perturbations in comparison with the competing methods. It demonstrated a stable performance across all metrics. Among the baselines, XE-MART\textsubscript{NDCG} that performed well in terms of NDCG@5, demonstrated poor performance in terms of AP@5. 

\paragraph{\textbf{Black-box Adversarial Attack:}} The black-box adversarial attack restricts the attacker’s knowledge only to the deployed model \cite{bhambri2019survey}. The setting of black-box attacks is closer to the real-world scenario; therefore, this is the most practical experiment to measure the robustness of our algorithm. Please refer to \cite{bhambri2019survey} for more information on the black-box adversarial attacks. Since the adversary has no access to the model's weights and parameters, the adversary can choose to train a parallel model called a \emph{substitute model} to imitate the original model. Usually, the attacker can use a much superior architecture than the original models for the weight estimation. Here, we use Neural Networks (NN) as our substitute models. The architecture of this model is presented in the supplement. To construct the substitute models, the training data were independently fed to each model and the output was observed. Then, for each algorithm, an NN ranking model was trained as an adversarial substitute model using the training feature vectors and the observed output of that specific algorithm. Subsequently, 75\% of the test queries were perturbed using the FGSM method and the parameters of the substitute model corresponding to each algorithm. 

\begin{figure}[!tb]
	\centering
	\subfloat[AP@5]{\includegraphics[width=0.23\textwidth]{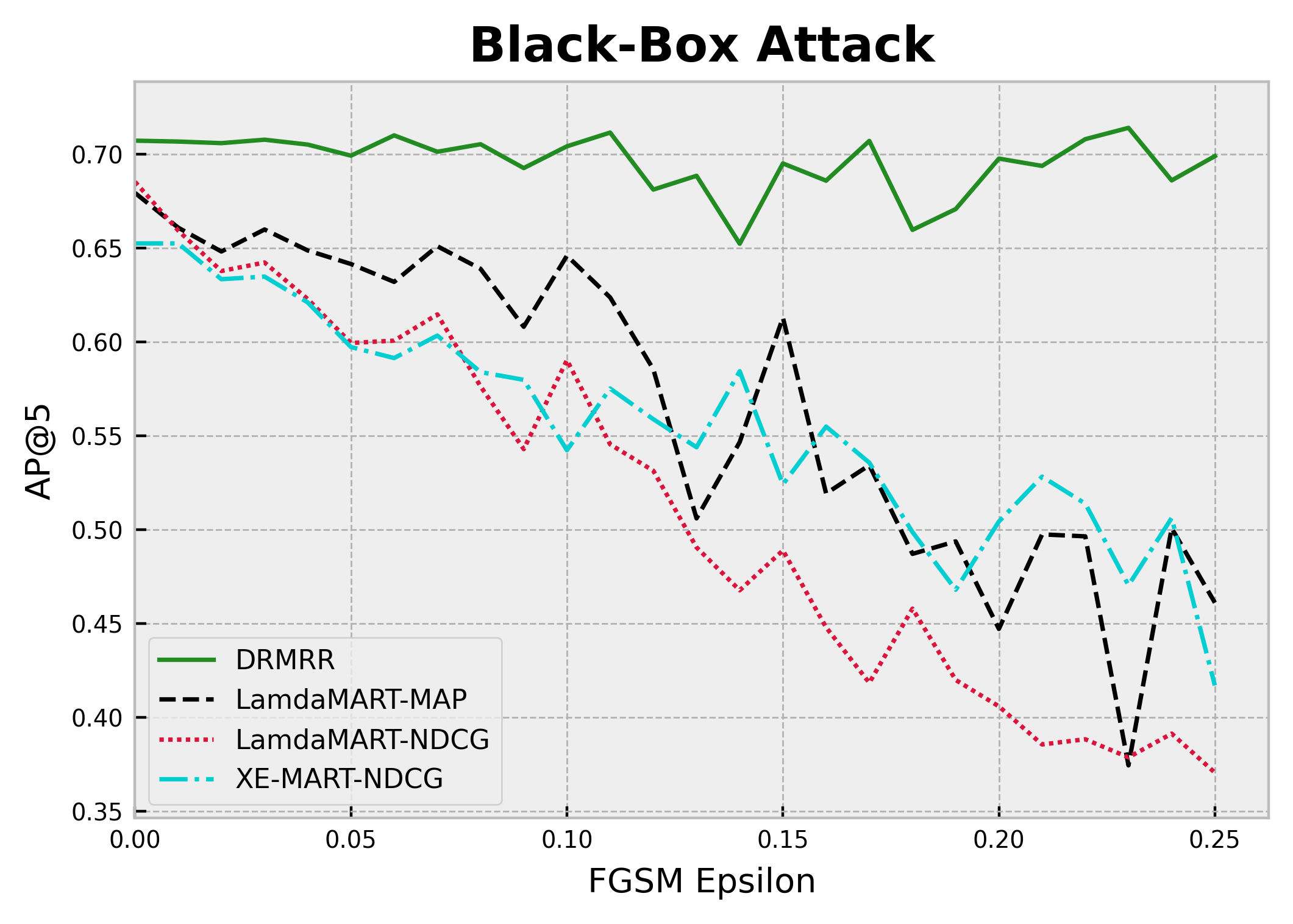}}\hfill
	\subfloat[NDCG@5]{\includegraphics[width=0.23\textwidth]{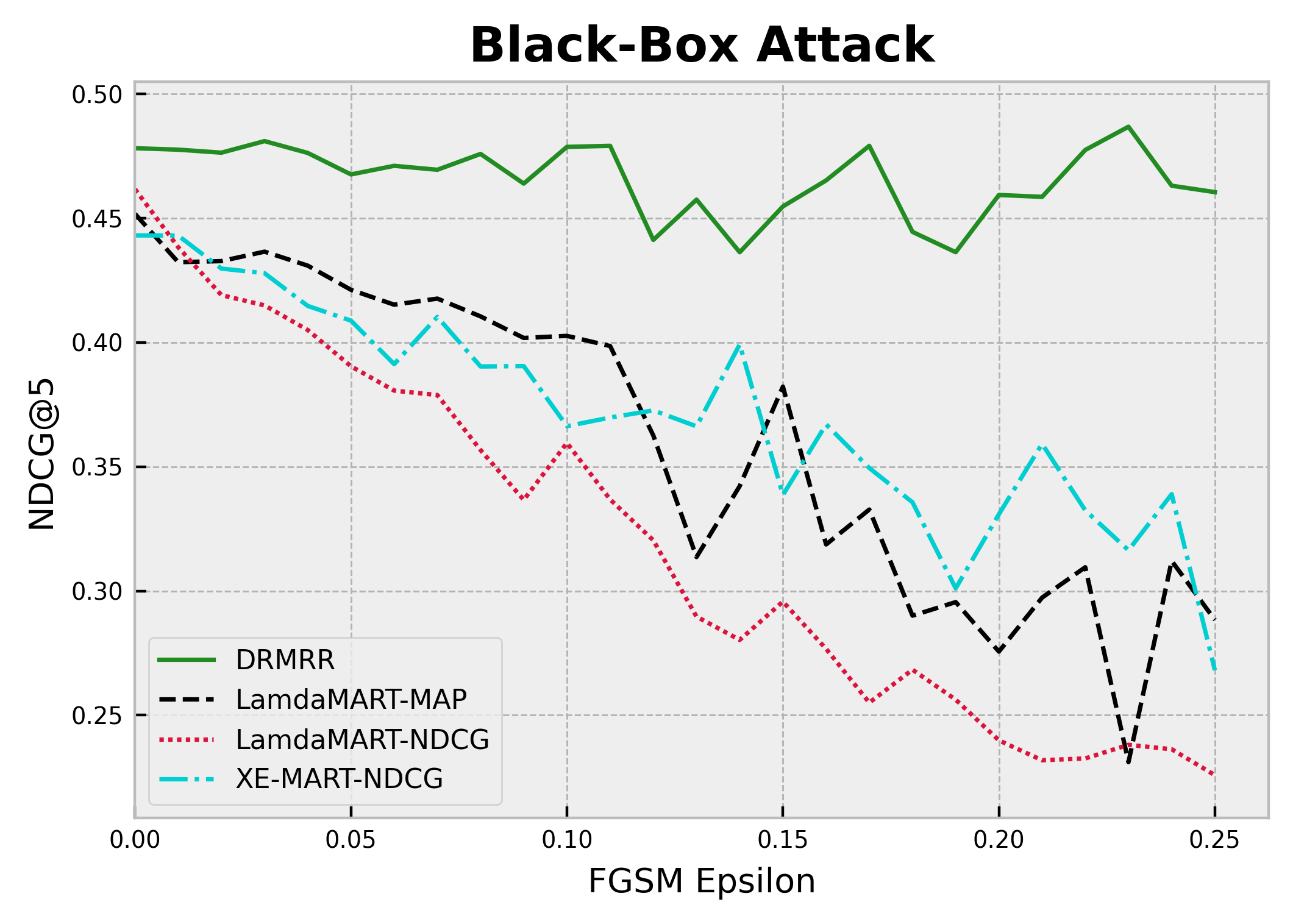}}	
	\caption{The impact of black-box adversarial perturbation on the performance of ranking models.}\label{fig:Robust_BB}
\end{figure}

We trained four substitute models corresponding to each ranking algorithm. We used the specific perturbed test data to evaluate the best trained models (i.e., each model has a different perturbed test set). Fig.~\ref{fig:Robust_BB} demonstrates the performance of the algorithms on the perturbed test data. The values are the average of 5 folds. We can see that both figures show the same trend -- increasing the level of perturbations (i.e., $\epsilon$) leads to significant differences between DRMRR and the baselines methods. The competing methods were greatly affected by this type of noise, whereas their performance was modest in the simpler experiments, namely universal adversarial attack and Gaussian attack. We conclude that DRMRR is robust to adversarial perturbations, an important property that leads to good generalization ability.

\paragraph{\textbf{Label Attack:}} In practice, the vagueness of query intent, insufficient domain knowledge, and ambiguous definition of relevance levels make it hard for human judges to assign proper relevance labels to some documents. Practically speaking, the probability of judgment errors in various relevance degrees is not equal. Even if human annotators misjudge a document, they are more probable to label it closer to its ground-truth label. Inspired by \cite{niu2015noise}, we define the non-uniform error probabilities in Table~\ref{table:tab2} where entries of this table correspond to the probability that a document with ground-truth label $y_{i}^{q}$ is prone to be labeled as $y_{j}^{q}$. We randomly changed the labels of the training data using the probabilities in Table \ref{table:tab2}. Then, each model was trained on the noisy training data. Clean test data were used to evaluate each model. We conducted two sets of experiment, namely low label noise (i.e., $e=0.85$) and high label noise (i.e., $e=0.7$). Table \ref{table:Robust_Label} reports the performance of the algorithms on the clean test data. The values in these figures are the average of 5 folds. For the low noise scenario, the differences between the average AP@5 and NDCG@5 of the baseline models and DRMRR were  2.53\% and 1.59\%, respectively. Notably, the gaps were even larger for the high noise scenario (AP@5=3.08\%, NDCG@5=1.68\%). Since noise in human-labeled data is an inevitable issue,  we can argue that the baseline models are susceptible and degrade more severely as more noise is added to the training set.

\begin{table}[!tb]
\centering
\resizebox{.9\columnwidth}{!}{
\begin{tabular}{ccccc}
\thickhline
\multirow{2}{*}{\textbf{$P(y_{i}^{q}\xrightarrow[]{}y_{j}^{q})$}} &                                 & \multicolumn{3}{c}{\textbf{$y_{j}^{q}$}}                                                               \\ \cline{3-5} 
                                                                   &                                 & \textbf{0}                              & \textbf{1}                              & \textbf{2}         \\ \thickhline
\multicolumn{1}{c|}{\multirow{3}{*}{\textbf{$y_{i}^{q}$}}}         & \multicolumn{1}{c|}{\textbf{0}} & \multicolumn{1}{c|}{$e$}                & \multicolumn{1}{c|}{$\frac{2}{3}(1-e)$} & $\frac{1}{3}(1-e)$ \\ \cline{3-5} 
\multicolumn{1}{c|}{}                                              & \multicolumn{1}{c|}{\textbf{1}} & \multicolumn{1}{c|}{$\frac{1}{2}(1-e)$} & \multicolumn{1}{c|}{$e$}                & $\frac{1}{2}(1-e)$ \\ \cline{3-5} 
\multicolumn{1}{c|}{}                                              & \multicolumn{1}{c|}{\textbf{2}} & \multicolumn{1}{c|}{$\frac{1}{3}(1-e)$} & \multicolumn{1}{c|}{$\frac{2}{3}(1-e)$} & $e$                \\ \thickhline
\end{tabular}}
\caption{The error probability table.}\label{table:tab2}
\end{table}


\begin{table}[!tb]
\centering
\resizebox{.98\columnwidth}{!}{
\begin{tabular}{ccccc}
\thickhline
\multicolumn{1}{l}{\textbf{}} & \multicolumn{2}{c}{\textbf{AP@5}}   & \multicolumn{2}{c}{\textbf{NDCG@5}} \\ \thickhline
\multicolumn{1}{l}{\textbf{}} & \textbf{High}    & \textbf{Low}     & \textbf{High}    & \textbf{Low}     \\ \thickhline
\textbf{DRMRR}                & \textbf{70.25\%} & \textbf{69.30\%} & \textbf{48.43\%} & \textbf{47.76\%} \\
\textbf{LamdaMART\textsubscript{MAP} }        & 67.97\%          & 67.87\%          & 47.16\%          & 45.87\%          \\
\textbf{LamdaMART\textsubscript{NDCG}}       & 67.82\%          & 66.79\%          & 46.89\%          & 46.46\%          \\
\textbf{XE-MART\textsubscript{NDCG}}         & 65.70\%          & 65.65\%          & 46.19\%          & 46.18\%          \\ \thickhline
\end{tabular}}
\caption{The impact of label noise on the performance of ranking models.}\label{table:Robust_Label}
\end{table}

\section{Discussion and Conclusion}
This paper went beyond conventional listwise learning-to-rank approaches and introduced a distributionally robust learning-to-rank framework with multiple outputs, referred to as DRMRR. Unlike existing methods, the scoring function in DRMRR was designed as a multivariate mapping from a feature vector to a vector of deviation scores (a.k.a. GTD vector). The GTD vector captures local context information and cross-document interactions. Moreover, we formulated DRMRR as a min-max problem where one minimizes a worst-case expected loss over a probabilistic ambiguity set. The ambiguity set was defined as a ball of distributions using the Wasserstein metric. Notably, we presented the compact and computationally solvable relaxations of the min-max formulation of DRMRR. We compared DRMRR with the baseline models in terms of $(a)$ the overall performance on two real-world applications and $(b)$ the robustness to various types and degrees of noise. In medical document retrieval, DRMRR outperformed state-of-the-art LTR models and established its capability in differentiating relevant documents from irrelevant ones.  In drug response prediction, our results indicated that DRMRR leads to substantially improved performance when compared to the competing methods across all performance metrics. Thus, DRMRR can infer robust predictors of drug responses from patient genomic or proteomic profiles which can lead to selecting highly effective personalized treatment.

In our robustness evaluations, we conducted a comprehensive analysis to assess the resilience of DRMRR against various types of noise and perturbations. Experimental results demonstrated that DRMRR is effective against: $(i)$ Gaussian noise; $(ii)$ universal adversarial perturbations by a substitute model with no knowledge of the victim model; $(iii)$ black-box adversarial perturbations by a substitute model with access only to the deployed victim model; and $(iv)$ probabilistic perturbation of relevance labels. Interestingly, the performance of DRMRR was consistently better than the baseline methods for all levels of noise. More importantly, DRMRR showed no significant change in its performance with the increase in the noise intensity.
Two attributes of DRMRR did help to enhance its performance and robustness: $(i)$ efficiently capturing the contextual information and interrelationship between documents/drugs via the GTD vector; and $(ii)$ the distributional robustness by hedging against a family of plausible distributions, including the true distribution with high confidence.

\bigskip

\end{document}